\pdfoutput=1

\documentclass[11pt]{article}

\usepackage[final]{acl}

\usepackage{times}
\usepackage{latexsym}

\usepackage[T1]{fontenc}

\usepackage[utf8]{inputenc}

\usepackage{microtype}

\usepackage{inconsolata}

\usepackage{graphicx}
\usepackage{subfigure}

\usepackage{amsmath}
\usepackage{amssymb}
\usepackage{mathtools}
\usepackage{amsthm}

\usepackage{hyperref}
\usepackage{url}

\usepackage{multirow}
\usepackage{color}
\usepackage{booktabs}
\usepackage{makecell}
\usepackage{tabularx}
\usepackage{fixltx2e}
\usepackage{tcolorbox}
\tcbuselibrary{breakable}

\title{Ada-Instruct: Adapting Instruction Generators for Complex Reasoning}

\author{Wanyun Cui \and Qianle Wang\\
Shanghai University of Finance and Economics\\
\texttt{cui.wanyun@sufe.edu.cn},
\texttt{wql20000111@stu.sufe.edu.cn}
}

\begin{document}
\maketitle

\begin{abstract}
Instructions augmentation is a crucial step for unleashing the full potential of large language models (LLMs) in downstream tasks. Existing Self-Instruct methods primarily simulate new instructions from a few initial instructions with in-context learning. However, our study identifies a critical flaw in this approach: even with GPT4o, Self-Instruct cannot generate complex instructions of length $\ge 100$, which is necessary in complex tasks such as code completion.

To address this issue, our key insight is that fine-tuning open source LLMs with only {\it ten examples} can produce complex instructions that maintain distributional consistency for complex reasoning tasks. We introduce Ada-Instruct, an \underline{ada}ptive \underline{instruct}ion generator developed through fine-tuning. We empirically validated Ada-Instruct's efficacy across different applications. The results highlight Ada-Instruct's  capacity to generate long, intricate, and distributionally consistent instructions.\footnote{Code is available at \url{https://github.com/wangitu/Ada-Instruct}}
\end{abstract}




\section{Introduction}

Supervised fine-tuning (SFT) is crucial for harnessing the potential of pre-trained Large Language Models (LLMs) in downstream tasks. Addressing SFT's demand for extensive training data, recent research has employed advanced LLMs, such as ChatGPT, to generate instructions. A prevalent approach is called ``Self-Instruct''~\citep{wang2022self}, which involves having ChatGPT sequentially generate both instructions and answers~\citep{sun2023principle, peng2023instruction, alpaca, schick2021generating, honovich2022unnatural, ye2022zerogen, meng2022generating, meng2023tuning}. It efficiently generates substantial novel training samples from a minimal number of initial samples. 


\begin{figure*}[tbh]
\centering
\subfigure[Self-Instruct with GPT-3.5-turbo-instruct on HumanEval.]{\label{fig:length:a0}
\includegraphics[width=0.28\textwidth]{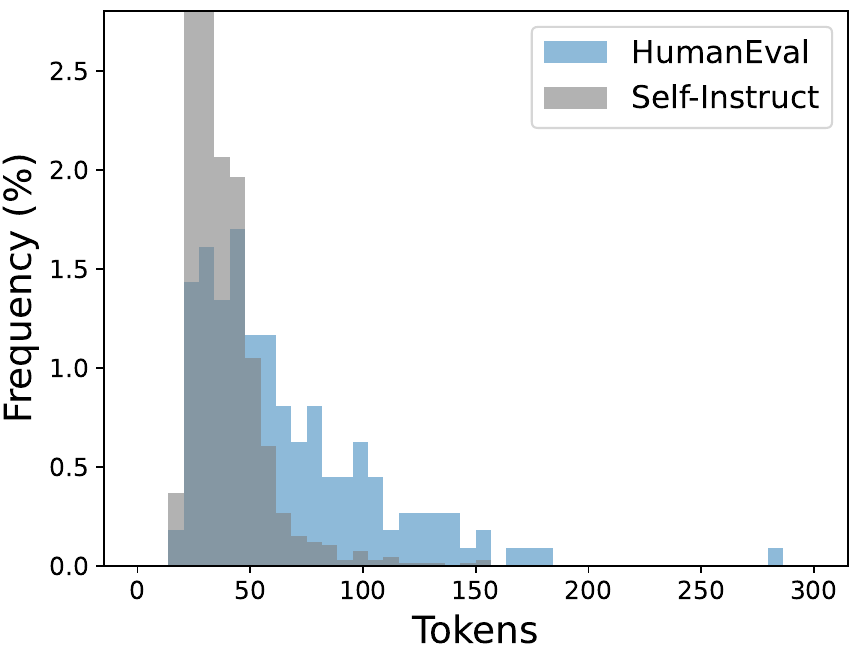}}
\hfill
\subfigure[Self-Instruct (prompt engineered) with GPT-3.5-turbo-instruct on HumanEval.]{\label{fig:length:a}
\includegraphics[width=0.28\textwidth]{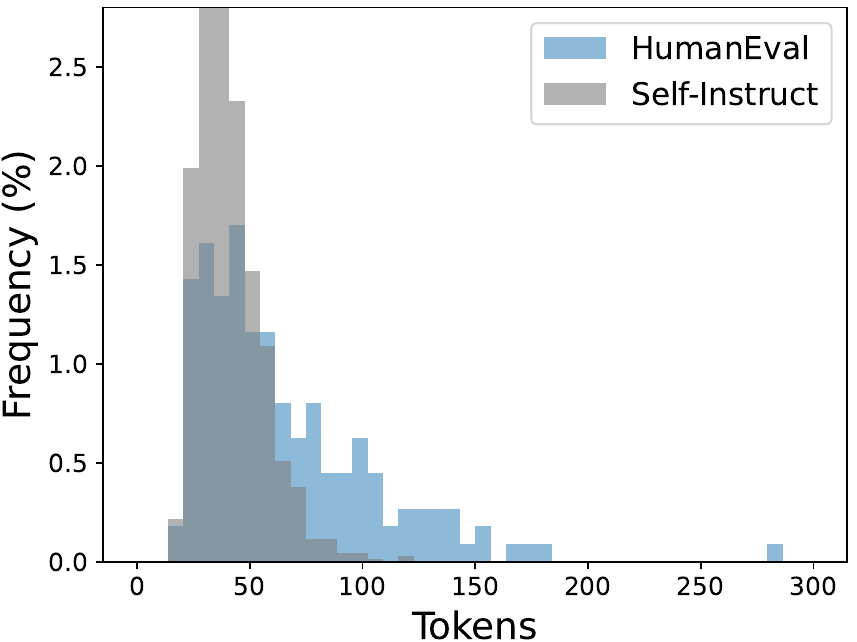}}
\hfill
\subfigure[Self-Instruct with GPT-4o on HumanEval.]{\label{fig:length:gpt4o:humaneval}
\includegraphics[width=0.28\textwidth]{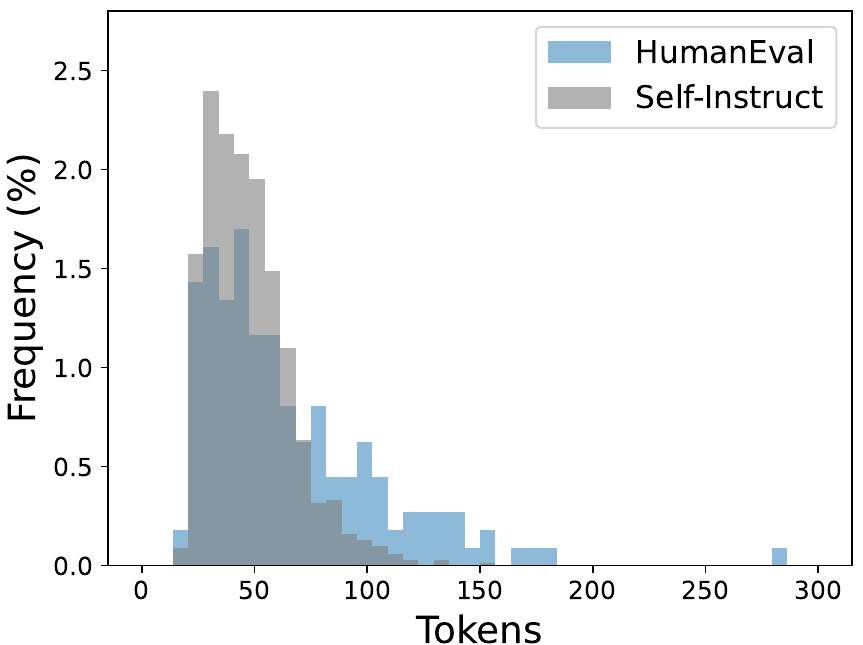}}
\subfigure[Self-Instruct with GPT-3.5-turbo-instruct on GSM8k.]{\label{fig:length:c0}
\includegraphics[width=0.28\textwidth]{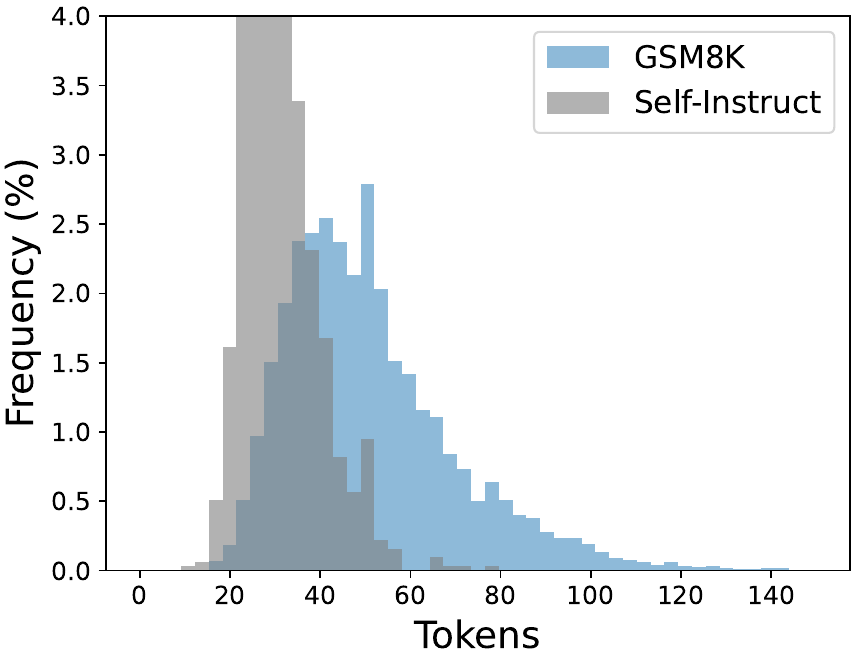}}
\hfill
\subfigure[Self-Instruct (prompt engineered) with GPT-3.5-turbo-instruct on GSM8k.]{\label{fig:length:c}
\includegraphics[width=0.28\textwidth]{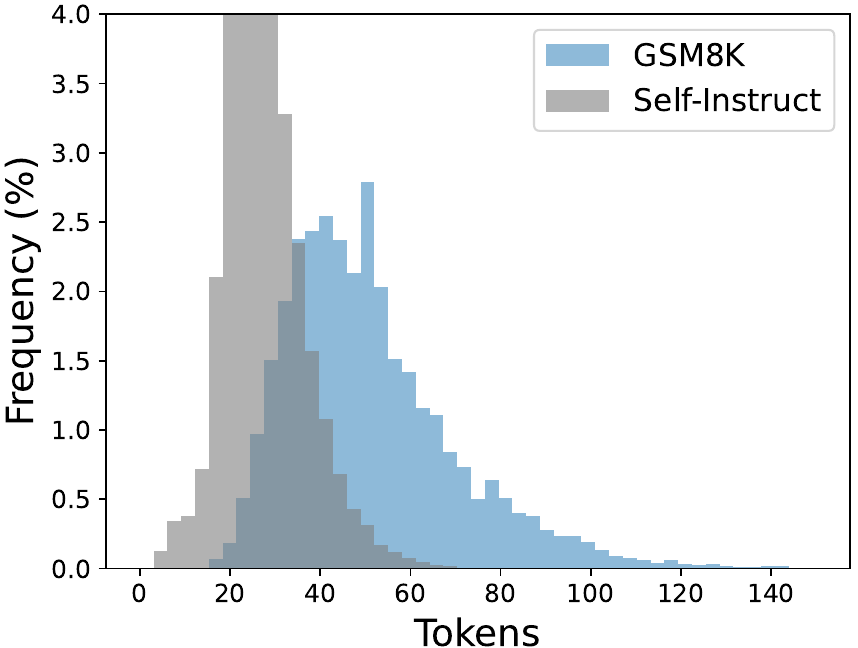}}
\hfill
\subfigure[Self-Instruct with GPT-4o on GSM8k.]{\label{fig:length:gpt4o:gsm8k}
\includegraphics[width=0.28\textwidth]{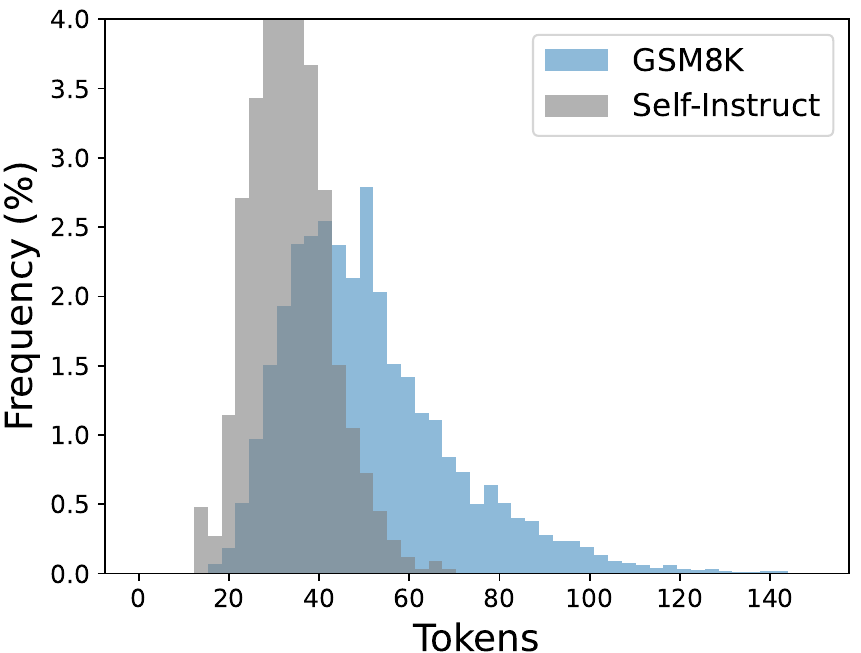}}
\subfigure[Ada-Instruct on HumanEval.]{\label{fig:length:b}
\includegraphics[width=0.28\textwidth]{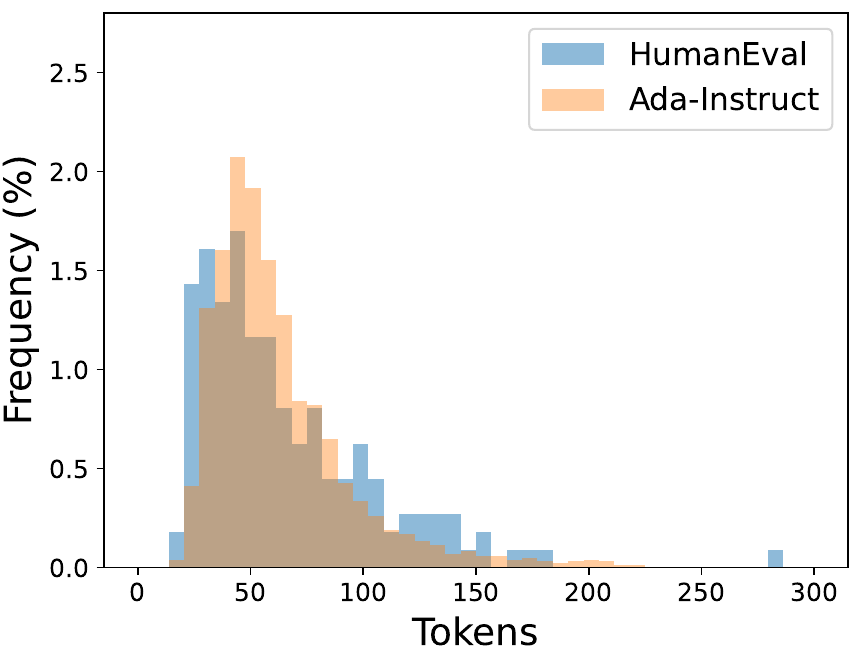}
}
\hspace{1cm}
\subfigure[Ada-Instruct on GSM8k.]{\label{fig:length:d}
\includegraphics[width=0.28\textwidth]{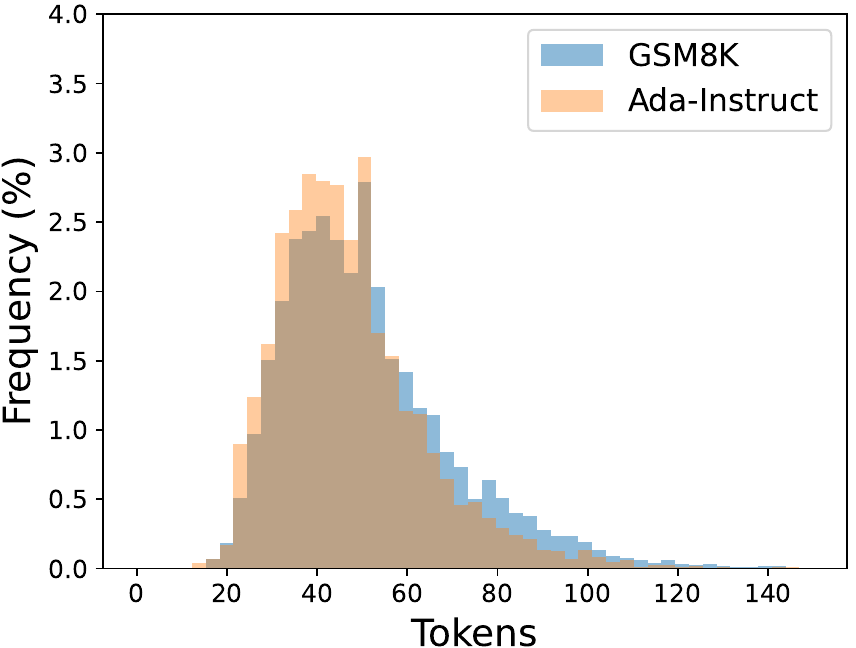}
}
\caption{Length Distribution of Different Methods. The length is measured by the number of tokens. All methods start with the same 10 instructions. \subref{fig:length:a0}\subref{fig:length:c0}: Self-Instruct struggles to generate complex instructions with more tokens, even being explicitly asked to do so~\subref{fig:length:a}\subref{fig:length:c}. \subref{fig:length:gpt4o:humaneval}\subref{fig:length:gpt4o:gsm8k}: The more advanced GPT-4o still has this issue. \subref{fig:length:b}\subref{fig:length:d}: Ada-Instruct successfully produces instructions whose length is consistently aligned with the target distribution.}
\label{fig:length}
\end{figure*}

However, our observations reveal a fundamental and critical limitation of Self-Instruct — {\it it notably struggles to generate complex instructions}. Despite being demonstrated with long and complex examples, Self-Instruct predominantly produces disappointingly brief and overly simplistic instructions. This is evident in Figure~\ref{fig:length:a0} and Figure~\ref{fig:length:c0}, where we present the length distribution of instructions by Self-Instruct on HumanEval (programming) and GSM8k (mathematics). The figures expose a glaring gap: Self-Instruct fails to produce instructions that exceed 100 and 60 tokens for HumanEval and GSM8k, respectively. 
This limitation significantly undermines the use of self-instruct in more complex tasks.

{\bf Is Prompt Engineering a Solution?} Despite its widespread use in enhancing in-context learning, prompt engineering is not the panacea it is often made out to be~\cite{wang2022self,sun2023principle,zhou2022large,yang2023large}. To encourage the generation of longer and more complex instructions, we explored infusing prompts with extra requirements, such as ``generate algorithms of intermediate level'' (for HumanEval) and ``the instructions should not be too easy'' (for GSM8k). However, as shown in Figure~\ref{fig:length:a}\ref{fig:length:c}, this approach did not effectively solve the problem of producing short instructions. A more advanced variant of prompt engineering, Evol-Instruct, employs multiturn strategies to incrementally enhance the complexity and variety of instructions. However, we will show in \S~\ref{sec:exp:creativity} that Evol-Instruct is unable to generate instructions that semantically align with the target instruction distribution.

{\bf Has the Problem Been Solved by the More Advanced GPT4o?} We performed additional evaluations using the advanced GPT4o model, which is equipped with superior reasoning and long-text processing capabilities. Figures~\ref{fig:length:gpt4o:humaneval}\ref{fig:length:gpt4o:gsm8k} illustrate that while GPT4o outperforms gpt-3.5-turbo-instruct in terms of average output length on the HumanEval benchmark, it still falls short of generating instructions longer than 100 tokens. Similarly, on the GSM8k benchmark, GPT4o shows no marked improvement in its capacity to produce longer instructions. Consequently, the challenge of generating complex instructions remains with the more advanced GPT4o.

In this paper, we unveil a novel insight into the instruction generation capabilities. Surprisingly, we find that {\it even when relying solely on 10 samples, a straightforward fine-tuned model is capable of generating instructions that align with the target task distribution.} In Figure~\ref{fig:length:b}, FT models generate instructions of length $\ge 100$ tokens for HumanEval, and in Figure~\ref{fig:length:d}, of length $\ge 60$ tokens for GSM8k, both matching the actual distribution. In addition, the generated instructions span the target distribution (\S~\ref{sec:exp:creativity}), and exhibit high diversity (\S~\ref{sec:exp:diversity}). 

Based on these findings, we introduce Ada-Instruct, a few-shot instruction generation procedure for downstream tasks. We fine-tune open-source LLMs using few-shot task samples for instruction generation, instead of ICL as in Self-Instruct. 

In summary, our contributions include (1) We uncover a new insight into the sample generation capabilities of self-instruct, showing that it cannot generate complex instructions. 
(2) We introduce Ada-Instruct, a few-shot instruction generation methodology with fine-tuning. (3)  We verify the effectiveness of Ada-Instruct through empirical validations, showcasing its superiority in generating complex instructions that are not only longer, but also aligned with the target distributions.

\begin{figure*}[t]
\centering
\includegraphics[width=1.0\textwidth]{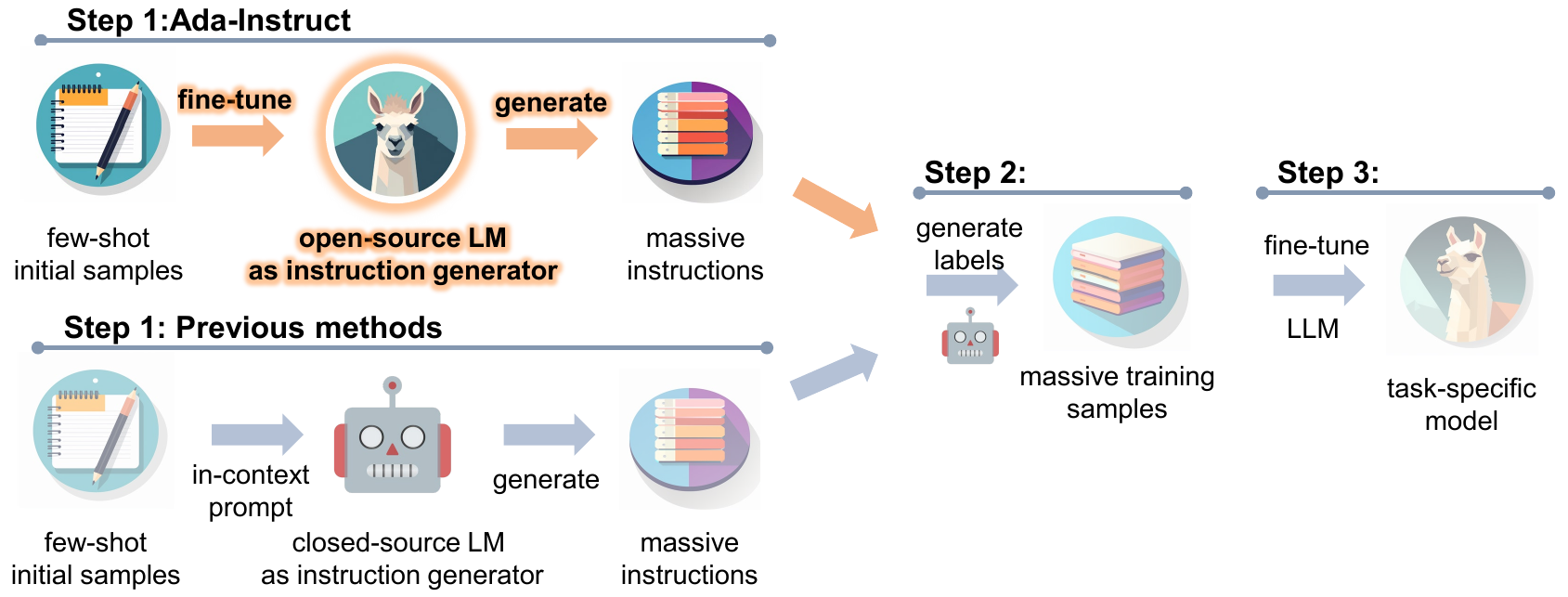}
\caption{How Ada-Instruct works. We fine-tune LLMs as instruction generators from few-shot initial samples (step 1), while previous self-instruct methods use in-context prompting and closed-source LLMs. We then use ChatGPT to generate labels (step 2), and fine-tune a task-specific model with the labeled samples (step 3).}
\label{fig:intro}
\end{figure*}

\section{Related Work}

{\bf Sample Generation via LLMs} Recent works have explored the use of LLMs for sample generation, often within the self-instruction framework~\citep{chen2023empirical}. This typically involves starting from an initial pool of instructions and having the LLMs iteratively generate new instructions along with the corresponding answers. Most prior work in the realm of instruction generation has relied on ICL~\citep{wang2022self,alpaca,sun2023principle,xu2023wizardlm,honovich2022unnatural,meng2022generating}. Various studies have focused mainly on improving the self-instruct approach in different problem scenarios. 

However, a limitation of this paradigm, as we have observed, is that ICL lacks the capacity to generate complex samples based solely on in-context examples. Although more intricate samples could potentially be produced using evolutionary strategies, such as Evol-Instruct~\citep{xu2023wizardlm,luo2023wizardmath,luo2023wizardcoder}, these manually designed tactics risk generating samples that do not align with the target task distribution.

FewGen~\citep{meng2023tuning} is the only method we have identified that substitutes fine-tuning for In-Context Learning (ICL) in sample generation. However, FewGen requires sophisticated metalearning and is limited to classification tasks. In contrast, Ada-Instruct is substantially simpler and more general. 

{\bf ICL vs. FT} Previous exploratory studies have aimed to compare the performance of ICL and FT methodologies. Some research suggests that ICL exhibits a more robust out-of-distribution generalization compared to FT~\citep{si2022prompting,awadalla2022exploring,utama2021avoiding}. However, some recent studies~\citep{mosbach2023few} argue that these earlier comparisons may be biased. The unfairness arises from using different model architectures for comparison (e.g., GPT-3-based ICL versus RoBERTa~\citep{liu2019roberta}-based FT) or by basing results on small-scale models. In more equitable experimental setups, the researchers found that FT outperforms ICL~\citep{mosbach2023few}, thereby supporting our strategy of using FT models for instruction generation.

\section{Method}


Ada-Instruct is divided into three steps: 1) Learning an instruction generator and generating massive instructions (\S~\ref{sec:method:step1}), 2) generating labels with ChatGPT (\S~\ref{sec:method:step2}), and 3) training LLMs for downstream tasks (\S~\ref{sec:method:step3}). In the following, we dive into the details of each step. The overall workflow is shown in Figure~\ref{fig:intro}.

\subsection{Learning to Generate Instructions (Step 1)}
\label{sec:method:step1}

The first step focuses on learning an instruction generator using a small set of samples. In most real-world scenarios, obtaining large labeled datasets for every new downstream task is infeasible. Hence, an instruction generator serves as an intermediary, converting small sets of samples into sufficient instructions for data labeling or task understanding.

Given a target downstream task \( T \) and a small set of samples \( S = \{ (x_1, y_1), (x_2, y_2), \ldots, (x_n, y_n) \} \), the objective is to fine-tune an initial LLM $M(\theta)$ with parameters $\theta$ to produce instructions \( I \) that have the same distribution as the instruction $X$ of task \( T \) and are beneficial for fine-tuning.

The goal of fine-tuning is learning to generate instructions $X$. Thus its objective is to optimize the parameters $\theta$ of the LLM to maximize the conditional likelihood of the target sequences given their corresponding instructions::
\begin{equation}
\mathcal{L}_{\text{inst}}(\theta) = -\frac{1}{n} \sum_{(x_i, y_i) \in S} \log P_M(x_i | \theta)
\end{equation}
Here, \( P_M(x_i | \theta) \) denotes the probability of observing the target instruction \( x_i \) under the current model parameters \( \theta \). $\theta$ is initialized as the pre-trained parameters. 
In causal language modeling, the probability of the target instruction is represented as the product of the conditional probabilities of the individual tokens in it.

\textbf{Generating Massive Instructions:}
After fine-tuning, the instruction generator is used to generate a large volume of instructions. The templates in this step are provided in Appendix \ref{sec:appendix:step1}. These instructions serve as the basis for the subsequent phases for generating high-quality samples.

\textbf{Filtering Duplicate Instructions:}
As massive instructions are generated from the LLM trained by a few samples, one issue is whether these instructions are duplicated. We assume that if two instructions are highly similar, using the two instructions to fine-tune the final LLM will be less effective. To further ensure the uniqueness of generated instructions, a simple filtering mechanism is used. This mechanism uses a pre-trained sentence embedding model to calculate the semantic similarity between generated instructions. If the semantic similarity between two instructions is above a predetermined threshold, the latter instruction is filtered out to avoid redundancy. In this paper, we use MPNet~\citep{song2020mpnet} to compute the semantic similarities.


\subsection{Label Generation (Step 2)}
\label{sec:method:step2}

In the second step, we leverage a high quality closed-source LLM, ChatGPT~\footnote{We use {\tt gpt-3.5-turbo-instruct} in this paper}, to generate labels for the instructions produced in step 1. Using ChatGPT alleviates the need for extensive manual labeling, providing a cost-efficient and time-effective way to accumulate labeled data based on the instructions generated in step 1~\citep{gilardi2023chatgpt}.

Given the set of instructions \( I = \{ x_1, x_2, \ldots, x_m \} \), the objective here is to generate their corresponding labels \(y_1, y_2, \ldots, y_m\). For each instruction \( I \) in the set, ChatGPT generates a corresponding response, transforming \( I \) into a new training set \( \mathbb{S} = \{(x_1,y_1), \ldots, (x_m,y_m) \} \).

\subsection{Training LLMs for Downstream Tasks (Step 3)}
\label{sec:method:step3}

The final step utilizes the complete training samples \( S' \) obtained from Step 2 to train LLMs for the target downstream tasks.

The objective function is also a casual language modeling loss over the given samples, adjusted to fit the labels of the new set of samples $\mathbb{S}$ from Step 2. A new LLM \( \mathbb{M}(\theta) \) is used for fine-tuning with the pre-trained parameter initialization:
\begin{equation}
\mathcal{L}_{\text{task}}(\theta) = -\frac{1}{m} \sum_{(x_i, y_i) \in \mathbb{S}} \log P_{\mathbb{M}}(y_i | x_i; \theta)
\end{equation}

\section{Experiments}

In our experiments, we evaluate the effectiveness of Ada-Instruct in code completion (\S~\ref{sec:exp:code}), mathematics (\S~\ref{sec:exp:math}), and commonsense reasoning (\S~\ref{sec:exp:commonsense}). We further analyze its distributional consistency with the target task, assessing (1) {\bf Semantic Consistency} (\S~\ref{sec:exp:creativity}): the alignment of generated examples with the target distribution, and (2) {\bf Diversity} (\S~\ref{sec:exp:diversity}): the variety in instructions from 10 initial samples. We also address the concern regarding whether fine-tuning an open-source model could result in diminished performance, considering that open-source models are often perceived as less qualified compared to closed-source models (\S~\ref{sec:exp:qualityaffect}). All experiments ran on a single node with 8 x A100 80GiB GPUs.

\subsection{Code Completion}
\label{sec:exp:code}

\begin{table}[tb]
\centering
\fontsize{8pt}{10pt}\selectfont
\setlength{\tabcolsep}{2pt}
\begin{tabularx}{\columnwidth}{lccc|c|c}
\toprule
\textbf{Model} &
  \textbf{\makecell[c]{Initial\\Data}} &
  \textbf{\makecell[c]{SFT\\Data}} &
  \textbf{Size} &
  \textbf{HumanEval} &
  \textbf{MBPP} \\ \midrule
Base model & - & - & \multicolumn{1}{c|}{13B} & 43.3 & 49.0 \\ \midrule
  \multicolumn{6}{c}{SOTA baselines}                               \\ \midrule
PaLM               & -   & -   & \multicolumn{1}{c|}{540B}  & \multicolumn{1}{c|}{26.2}          & 36.8 \\
PaLM-Coder         & -   & -   & \multicolumn{1}{c|}{540B}  & \multicolumn{1}{c|}{36.0}          & 47.0 \\
PaLM 2-S           & -   & -   & \multicolumn{1}{c|}{-}     & \multicolumn{1}{c|}{37.6}          & 50.0 \\
StarCoder\textsubscript{Python}   & -   & -   & \multicolumn{1}{c|}{15.5B} & \multicolumn{1}{c|}{33.6}          & 52.7 \\
StarCoder\textsubscript{Prompted} & -   & -   & \multicolumn{1}{c|}{15.5B} & \multicolumn{1}{c|}{40.8}          & 49.5 \\
Code-Cushman\textsubscript{001}   & -   & -   & \multicolumn{1}{c|}{12B}   & \multicolumn{1}{c|}{33.5}          & 45.9 \\
GPT-3.5            & -   & -   & \multicolumn{1}{c|}{-}     & \multicolumn{1}{c|}{48.1}          & 52.2 \\
GPT-4              & -   & -   & \multicolumn{1}{c|}{-}     & \multicolumn{1}{c|}{\textbf{67.0}} & -    \\ \midrule
\multicolumn{6}{c}{Self-Instruct baselines}                                   \\ \midrule
Self-Instruct\textsubscript{HE} & 10 & 10k & \multicolumn{1}{c|}{13B} & 47.0 \textsubscript{(+8.5\%)} & - \\ 
Self-Instruct\textsubscript{MBPP} & 10 & 10k & \multicolumn{1}{c|}{13B} & - & 51.2 \textsubscript{(+4.5\%)}\\ 
Evol-Instruct        & 20k & 78k & \multicolumn{1}{c|}{13B}   & \multicolumn{1}{c|}{64.0\textsubscript{(+47.8\%)}}          & 55.6\textsubscript{(+13.5\%)} \\ 
\midrule
\textbf{Ada-Instruct\textsubscript{HE}} &
  10 &
  10k &
  \multicolumn{1}{c|}{13B} &
  \multicolumn{1}{c|}{\textbf{65.2} \textsubscript{(\textcolor{red}{+50.6\%})}} &
  - \\
\textbf{Ada-Instruct\textsubscript{MBPP}} &
  10 &
  10k &
  \multicolumn{1}{c|}{13B} &
  \multicolumn{1}{c|}{-} &
  \textbf{55.6} \textsubscript{(\textcolor{red}{+13.5\%})} \\
\bottomrule
\end{tabularx}
\caption{Results of pass@1 (\%) on HumanEval and MBPP, showcasing relative improvements over the base model. Results related to Code LLAMA are from~\cite{roziere2023code}. Results of other baselines and from~\cite{luo2023wizardcoder}. We follow \cite{roziere2023code} to adopt a greedy decoding strategy in Ada-Instruct.}
\label{tab:exp:code}
\end{table}

\textbf{Setup:}
We utilize two widely recognized benchmarks: HumanEval~\citep{chen2021evaluating} and MBPP~\citep{austin2021program}. For both benchmarks, our experiments began with an initial set of 10 samples. Specifically for MBPP, these samples were randomly extracted from its development set. For HumanEval, which does not have a development set, we selected 10 representative problems from LeetCode and the MBPP development set. This selection was aimed at closely mirroring the difficulty level as in HumanEval. These chosen samples were then appropriately formatted to align with HumanEval's query structure. We developed two models based on the instructions generated for HumanEval and MBPP, named Ada-Instruct\textsubscript{HE} and Ada-Instruct\textsubscript{MBPP}, respectively. We use Code LLAMA-Python (13B)~\citep{roziere2023code} as our base model. 

\textbf{Baselines:} The primary baseline is Self-Instruct. We ensure that it utilized an identical set of initial samples and the same quantity of SFT samples for a fair comparison. We denote two models built on the two generated instruction sets as Self-Instruct\textsubscript{HE} and Self-Instruct\textsubscript{MBPP}, respectively.
Another vital baseline was Evol-Instruct (WizardCoder~\citep{luo2023wizardcoder}), selected to evaluate the impact of sophisticated multi-turn prompt engineering techniques. We use the WizardCoder-Python-13B version, which also uses Code LLAMA-Python (13B) as the base model. Furthermore, our analysis included comparisons with leading-edge models in the field, such as PaLM~\citep{chowdhery2022palm}, PaLM-Coder~\citep{chowdhery2022palm}, PaLM 2-S~\citep{anil2023palm}, StarCoder~\citep{li2023starcoder}, and GPTs~\citep{openai2023gpt}, to establish a comprehensive comparison with the current state-of-the-art. 

\textbf{Main Results: Effect of Ada-Instruct:}
We show the results in Table~\ref{tab:exp:code}. Compared to state-of-the-art baselines, Ada-Instruct maintains a significant advantage in effectiveness.
Its pass@1 rate is second only to GPT-4. Compared to the base model (Code LLAMA-Python), Ada-Instruct exhibits a notable improvement in performance. This enhancement is particularly significant on HumanEval, where the relative increase reaches 50.6\%, even when initiated with as few as 10 samples. This substantial boost underscores the adaptability of Ada-Instruct, illustrating its ability to adapt LLMs to downstream tasks. The results lend evidence to Ada-Instruct's efficacy in optimizing language models for specific tasks.

\textbf{Comparison with Self-Instruct} 
We compared the performance of Ada-Instruct with Self-Instruct baselines. It is clear that with the same initial samples and the same amount of SFT data, Ada-Instruct significantly surpasses Self-Instruct in effectiveness. Ada-Instruct also shows superior performance compared to WizardCoder, which uses multi-turn prompting. Notably, WizardCoder requires 20k initial samples and 78k SFT data, which is considerably more than the sample size used by Ada-Instruct. These comparisons validate the superiority of Ada-Instruct over Self-Instruct in terms of effectiveness. We will further elaborate in Sec~\ref{sec:exp:analysis} that the instructions generated by Ada-Instruct exhibit greater semantic consistency, diversity, and coverage compared to those produced by Self-Instruct and Evol-Instruct.




\textbf{Generalization Abilities for Multiple Tasks}
To validate its generalization ability, we also adapt Ada-Instruct to target a domain of multiple tasks rather than a single task. This is achieved by expanding the initial sample pool to include initial samples from different tasks. We conducted a direct experiment: We used an initial sample set comprising 10 initial HumanEval samples and 10 initial MBPP samples. Using these 20 initial samples, our Ada-Instruct framework generated 10k instructions in total. We then trained a domain model, termed Ada-Instruct\textsubscript{Program}. For comparison, we also tested the performance of Self-Instruct using the same 20 initial samples and the same amount of SFT samples, denoted as Self-Instruct\textsubscript{Program}. As shown in Table~\ref{tab:multiple_tasks}, it is evident that Ada-Instruct still achieves a significant performance improvement in the target domain with just 20 initial samples, surpassing the results of Self-Instruct.

\begin{table}[tb]
\centering
\fontsize{9pt}{12pt}\selectfont
\setlength{\tabcolsep}{2pt}
\begin{tabularx}{\columnwidth}{lcc|cc}
\toprule
\textbf{Model}             & \textbf{\makecell[c]{Initial\\Data}} & \textbf{\makecell[c]{SFT\\Data}} & \textbf{HumanEval} & \textbf{MBPP} \\ \midrule
Base model & - & - & 43.3 & 49.0 \\
Self-Instruct\textsubscript{Program}      & 20                    & 10k               & 51.8\textsubscript{(+19.6\%)}               & 47.8\textsubscript{(-2.5\%)}          \\ 
Ada-Instruct\textsubscript{Program}       & 20                    & 10k               & {\bf 62.8}\textsubscript{(\textcolor{red}{+45.0\%})}               & {\bf 54.0}\textsubscript{(\textcolor{red}{+10.2\%})} \\ \bottomrule
\end{tabularx}
\caption{Results of pass@1 (\%) on multiple code completion tasks.}
\label{tab:multiple_tasks}
\end{table}

{\bf Effect on Unseen Tasks}
We also assessed the generalization capability on unseen tasks within the code completion domain. Specifically, we tested two scenarios:
\begin{enumerate}
    \item Utilize 10 initial HumanEval instructions and generate 10k SFT instructions. Then evaluate the fine-tuned model on MBPP.
    \item Utilize 10 initial MBPP instructions and generate 10k SFT instructions. Then evaluate the fine-tuned model on HumanEval.
\end{enumerate}

\begin{table}[tb]
    \centering
    \fontsize{9pt}{12pt}\selectfont
    \setlength{\tabcolsep}{3.5pt}
    \begin{tabularx}{\columnwidth}{lcc|c}
    \toprule
    \textbf{Model} &
    \textbf{\makecell[c]{Training\\Data}} & \textbf{\makecell[c]{Evaluation\\Task}} & \textbf{Pass@1} \\ \midrule
    Base model & - & HumanEval & 43.3 \\
    Self-Instruct & 10k HumanEval & HumanEval & 47.0 \\
    \textbf{Ada-Instruct} & \textbf{10k MBPP} & \textbf{HumanEval} & \textbf{60.4} \\ \midrule
    Base model & - & MBPP & 49.0 \\
    Self-Instruct & 10k MBPP & MBPP & 51.2 \\
    \textbf{Ada-Instruct} & \textbf{10k HumanEval} & \textbf{MBPP} & \textbf{52.4} \\
    \bottomrule
    \end{tabularx}
    \caption{Results of pass@1 (\%) on unseen code completion tasks.}
    \label{tab:unseen_tasks}
\end{table}

As presented in Table~\ref{tab:unseen_tasks}, Ada-Instruct demonstrates robust generalization abilities on unseen tasks, even outperforms self-instruct which was trained on the target task.

\subsection{Math}
\label{sec:exp:math}

\textbf{Setup:} We evaluated Ada-Instruct on two benchmarks: GSM8k~\citep{cobbe2021training} (easier) and MATH~\citep{hendrycks2021measuring} (harder). We randomly sampled 10 instructions from the training set of each benchmark as the initial samples. We require that the 10 MATH samples not be related to drawing scripts. We developed two models based on the instructions generated for each benchmark, named Ada-Instruct\textsubscript{GSM8k} and Ada-Instruct\textsubscript{MATH}, respectively. 
The base model used here was LLAMA 2.

\textbf{Baselines:} We employed Self-Instruct as the baseline. The models developed using initial instructions from GSM8k and MATH are respectively denoted as Self-Instruct\textsubscript{GSM8k} and Self-Instruct\textsubscript{MATH}. We have omitted the comparison with Evol-Instruct, as its implementation in WizardMath~\cite{luo2023wizardmath} already incorporates GSM8k and MATH as part of their training datasets. 

\begin{table}[h]
\centering
\fontsize{8pt}{10pt}\selectfont
\setlength{\tabcolsep}{2pt}
\begin{tabularx}{\columnwidth}{lccccc}
\toprule
\textbf{Model} &
  \textbf{\makecell[c]{Initial\\Data}} &
  \textbf{\makecell[c]{SFT\\Data}} &
  \textbf{Size} &
  \textbf{GSM8k} &
  \textbf{MATH} \\ \midrule
Base model & 8 & - & 13B & 28.7 & 3.9 \\ \midrule
\multicolumn{6}{c}{SOTA Models}                                                                 \\ \midrule                 
Falcon           & -   & -   & \multicolumn{1}{c|}{40B}  & \multicolumn{1}{c|}{19.6} & 2.5  \\
Baichuan-chat    & -   & -   & \multicolumn{1}{c|}{13B}  & \multicolumn{1}{c|}{23.9} & -      \\
Vicuna v1.3      & -   & -   & \multicolumn{1}{c|}{13B}  & \multicolumn{1}{c|}{27.6} & -      \\
GPT3             & -   & -   & \multicolumn{1}{c|}{175B} & \multicolumn{1}{c|}{34.0} & 5.2  \\
Text-davinci-002 & -   & -   & \multicolumn{1}{c|}{175B} & \multicolumn{1}{c|}{40.7} & 19.1 \\
Chinchilla       & -   & -   & \multicolumn{1}{c|}{70B}  & \multicolumn{1}{c|}{43.7} & -      \\
LLAMA 2          & -   & -   & \multicolumn{1}{c|}{34B}    & \multicolumn{1}{c|}{42.2} & 6.2      \\
LLAMA 2          & -   & -   & \multicolumn{1}{c|}{70B}    & \multicolumn{1}{c|}{56.8} & 13.5      \\
GPT-3.5          & -   & -   & \multicolumn{1}{c|}{-}    & \multicolumn{1}{c|}{57.1} & -      \\
PaLM 2           & -   & -   & \multicolumn{1}{c|}{540B} & \multicolumn{1}{c|}{80.7} & 34.3 \\
GPT-4 &
  - &
  - &
  \multicolumn{1}{c|}{-} &
  \multicolumn{1}{c|}{\textbf{92.0}} &
  \textbf{42.5} \\ \midrule
\multicolumn{6}{c}{Self-Instruct Baselines}                           \\ \midrule
Self-Instruct\textsubscript{GSM8k}       & 10 & 10k & \multicolumn{1}{c|}{13B}  & \multicolumn{1}{c|}{30.8 \textsubscript{(+7.3\%)}} & \multicolumn{1}{c}{-} \\
Self-Instruct\textsubscript{MATH}       & 10 & 10k & \multicolumn{1}{c|}{13B}  & \multicolumn{1}{c|}{-} & \multicolumn{1}{c}{5.8 \textsubscript{(+48.7\%)}} \\ \midrule
\textbf{Ada-Instruct\textsubscript{GSM8k}} &
  10 &
  10k &
  \multicolumn{1}{c|}{13B} &
  \multicolumn{1}{c|}{\textbf{48.7} \textsubscript{(\textcolor{red}{+69.7\%})}} &
  - \\
\textbf{Ada-Instruct\textsubscript{MATH}} &
  10 &
  10k &
  \multicolumn{1}{c|}{13B} &
  \multicolumn{1}{c|}{-} &
  \textbf{8.8} \textsubscript{(\textcolor{red}{+125.6\%})} \\
\bottomrule
\end{tabularx}
\caption{Results on GSM8k and MATH, demonstrating relative improvements over the base model (LLAMA 2). For the base model, we follow~\cite{touvron2023llama} to deploy 8-shot in-context learning. Results of baselines are from~\cite{luo2023wizardmath}. The decoding strategy of Ada-Instruct was sourced from~\cite{luo2023wizardmath}.}
\label{tab:math}
\end{table}

\textbf{Effect:} In Table~\ref{tab:math}, we observed a significant performance enhancement of Ada-Instruct in comparison with the base model. Ada-Instruct demonstrated a relative improvement of 69.7\% and 125.6\% on GSM8k and MATH, respectively, compared to the base model (LLAMA 2-13B). This surpassed the performance of LLAMA 2-34B and achieved state-of-the-art results in few-shot instruction generation models. 

\textbf{Comparison with Self-Instruct:} In Table~\ref{tab:math}, we also compare the performance of Ada-Instruct and Self-Instruct. The settings for both Self-Instruct and Ada-Instruct are kept consistent. 
Ada-Instruct markedly surpasses Self-Instruct. 

\subsection{Commonsense Reasoning}
\label{sec:exp:commonsense}
{\bf Setup:} We evaluated the effectiveness of Ada-Instruct on CommonsenseQA~\citep{talmor2019commonsenseqa}, a benchmark for commonsense reasoning. We randomly selected 10 samples from the training set to serve as initial samples. We choose LLAMA 2-13B as our base model. 

{\bf Baselines:} We compare with Self-Instruct with the same initial samples and the same amount of SFT data. We also compare with Evol-Instruct with the implementation of WizardLM~\cite{xu2023wizardlm}). For a fair comparison, we used the WizardLM-13B-V1.2 version, which also employs LLAMA2-13B as its base model.

{\bf Results:} Based on the results presented in Table~\ref{tab:csqa}, we observe a substantial improvement in performance attributed to Ada-Instruct. 
Ada-Instruct also demonstrated superior performance compared to both Self-Instruct and Evol-Instruct.

\begin{table}[tb]
\centering
\fontsize{8pt}{10pt}\selectfont
\begin{tabularx}{\columnwidth}{lcccc}
\toprule
\textbf{Model}  & \textbf{\makecell[c]{Initial\\Data}} & \textbf{\makecell[c]{SFT\\Data}}  & \textbf{Size}                           & \textbf{Accuracy}  \\ 
\midrule
\multicolumn{5}{c}{Base Models}                                                                         \\ \midrule
LLAMA 2 (0-shot)        & -   & -             & \multicolumn{1}{c|}{13B} & 59.0*                   \\
LLAMA 2 (1-shot)        & 1   & -             & \multicolumn{1}{c|}{13B}                     & 62.8*                   \\
\multirow{1}{*}{LLAMA 2 (7-shot)}  & 7   & -  & \multicolumn{1}{c|}{13B}                  & 67.3                    \\
LLAMA 2 (10-shot)       & 10   & -             & \multicolumn{1}{c|}{13B}                     & 68.1*                   \\ \midrule
\multicolumn{5}{c}{SOTA Models}     \\ \midrule
GPT-NeoX            & -   & -                & \multicolumn{1}{c|}{20B}                  & 60.4                    \\
BLOOM                & -   & -               & \multicolumn{1}{c|}{176B}                 & 64.2                    \\
OPT                  & -   & -               & \multicolumn{1}{c|}{66B}                  & 66.4                    \\
BloombergGPT         & -   & -               & \multicolumn{1}{c|}{51B}                  & 65.5                    \\ 
ChatGPT  & -   & -  & \multicolumn{1}{c|}{-} & 74.0 \\ \midrule
\multicolumn{5}{c}{Self-Instruct Baselines}                                                                         \\ \midrule

Self-Instruct  & 10   & 10k  & \multicolumn{1}{c|}{13B} & 71.4*\textsubscript{(+21.0\%)} \\ 
Evol-Instruct  & 52k   & 250k  & \multicolumn{1}{c|}{13B} & 64.0*\textsubscript{(+8.5\%)} \\ \midrule
\textbf{Ada-Instruct}    & 10   & 10k               & \multicolumn{1}{c|}{13B}                  & \textbf{75.5*} \textsubscript{(\textcolor{red}{+28.0\%})} \\
\bottomrule
\end{tabularx}
\caption{Results on CommonsenseQA. Results related to LLAMA 2 are from~\cite{touvron2023llama}. Results of other baselines are from~\cite{wu2023bloomberggpt}. *: results are tested on the dev set.}
\label{tab:csqa}
\end{table}

\subsection{Analysis of Distributional Consistency}
\label{sec:exp:analysis}


We have already illustrated in Figure~\ref{fig:length} that Ada-Instruct is capable of generating instructions whose length distribution aligns with the target task. We will now proceed to further analyze their semantic consistency. Given that we only used 10 initial samples, our investigation particularly focuses on two critical concerns: (1) the extent to which the generated instructions encompass the entire distribution of the target task, rather than merely echoing these initial examples (\S~\ref{sec:exp:creativity}), and (2) the diversity of the generated instructions, specifically examining whether they demonstrate a broad spectrum of variation 
(\S~\ref{sec:exp:diversity}).

\subsubsection{Semantic Distribution}
\label{sec:exp:creativity}

\begin{figure}[tb]
\centering
\subfigure[Semantic distribution of MBPP]{\label{fig:distribution:a}
\includegraphics[width=0.45\textwidth]{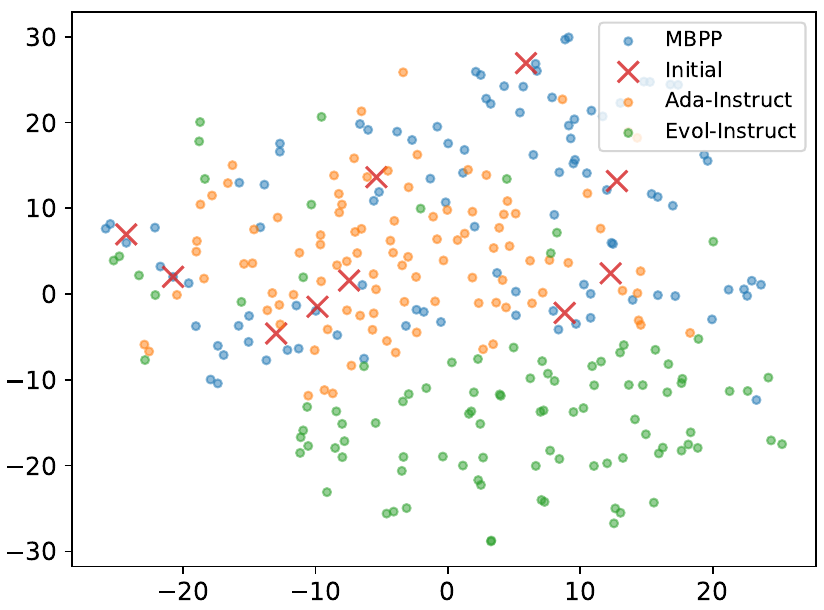}}
\subfigure[Semantic distribution of HumanEval]{\label{fig:distribution:b}
\includegraphics[width=0.45\textwidth]{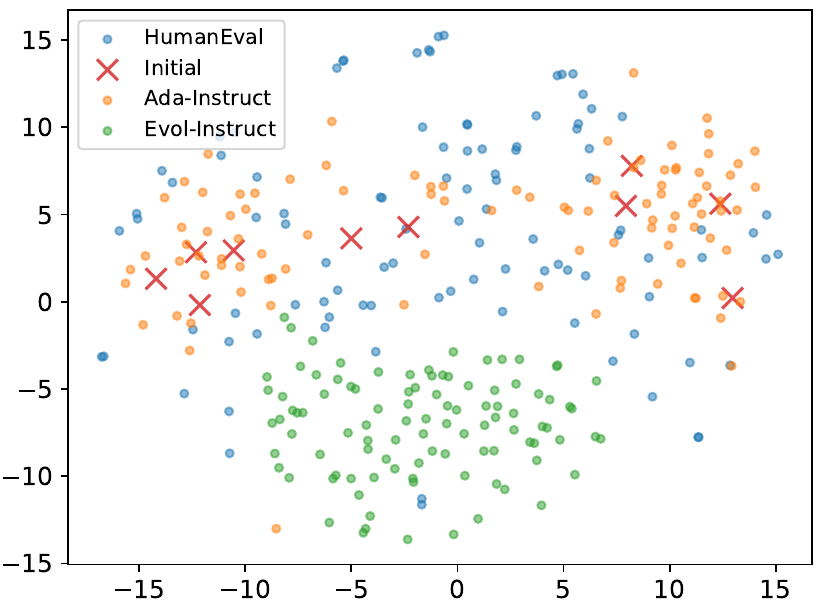}
}
\caption{Semantic distribution of generated instructions by t-SNE. Ada-Instruct shows better semantic distribution consistency than Evol-Instruct.}
\label{fig:distribution}
\end{figure}

We plot the semantic distribution of the initial instructions and the generated instructions. Additionally, we plot the distribution of the target task for comparison, to verify whether the generated instructions align with the target distribution. For comparison, we also plot the distribution of instructions by Evol-Instruct. 
We represent the semantics of the instructions using 
{\it text-embedding-ada-002} API from OpenAI
and visualized their distribution using t-SNE~\citep{van2008visualizing}.


Figure~\ref{fig:distribution} shows that the generated instructions exhibit a consistent distribution with the target task. The instructions of Ada-Instruct are not confined to the vicinity of the ten initial samples but demonstrate the capability to expand to broader regions, aligning with the actual instruction distribution of the target task. In contrast, the Evol-Instruct distribution shows noticeable deviations from the target instruction distribution. Such gaps are not unusual - Evol-Instruct, which is based on multi-turn prompt engineering, can generate long and complex instructions. However, crafting prompts manually without learning makes it difficult to fit the intended distribution. Ada-Instruct is capable of learning to adapt to the downstream instruction distribution. which is essential for instruction generation. These observations validate both Ada-Instruct's distributional consistency with respect to semantics, and the motivation of adapting LLMs as instruction generators for intended tasks.

\subsubsection{Diversity}
\label{sec:exp:diversity}

Given that our instruction generator was trained from merely 10 examples, another concern is whether the generated instructions are sufficiently diverse or if they overfit to a limited number of training samples. To address this, we assessed the diversity of the generated samples. Specifically, we randomly sampled 10000 pairs of generated samples for MBPP and calculated their similarity scores. A high similarity score for a pair of instructions indicates redundancy. Therefore, for a more diverse set of generated samples, we desire a lower similarity score distribution. We compared the diversity of instructions generated by Ada-Instruct and by Self-Instruct.

\begin{figure}[tb]
\centering
\includegraphics[width=0.45\textwidth]{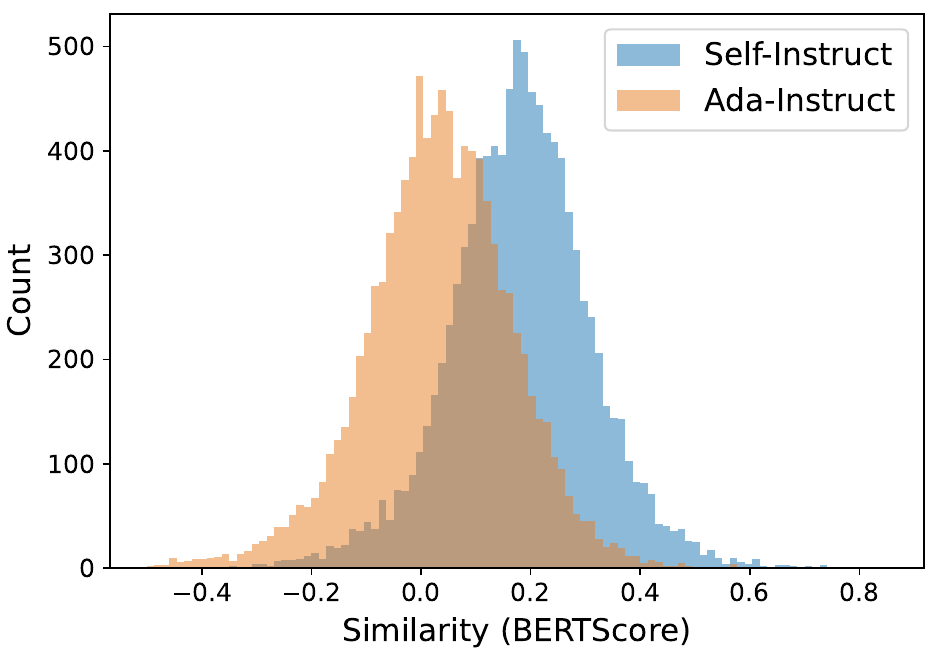}
\caption{Similarity score distribution. Ada-Instruct generally has lower similarity scores than Self-Instruct, indicating that it has high diversity.}
\label{exp:diversity}
\end{figure}

We followed the approach used in a previous work~\citep{honovich2022unnatural} to employ BERTscore~\citep{zhang2019bertscore} to measure the similarity between instruction pairs. The visualization of the results can be seen in Figure~\ref{exp:diversity}. The samples from Ada-Instruct exhibited lower similarity between pairs. This indicates that Ada-Instruct produces instructions with greater diversity. Given that the expressive capacity of the base model for Ada-Instruct (Code LLAMA) is evidently weaker than that of ChatGPT, this underscores the effectiveness of Ada-Instruct in generating diverse instructions. 

\subsection{The Impact of Instruction Quality}
\label{sec:exp:qualityaffect}

Ada-Instruct typically employs fine-tuning on open-source models, whereas Self-Instruct often uses closed-source models (like ChatGPT) for generating instructions. It is important to note that, as of now, the quality of open-source models generally lags behind that of closed-source models. Therefore, a concern with Ada-Instruct is that the quality of individual instructions might be lower, particularly for complex tasks. In this subsection, we investigate the actual impact on instruction quality.

\begin{figure}[t]
\centering
\includegraphics[width=0.42\textwidth]{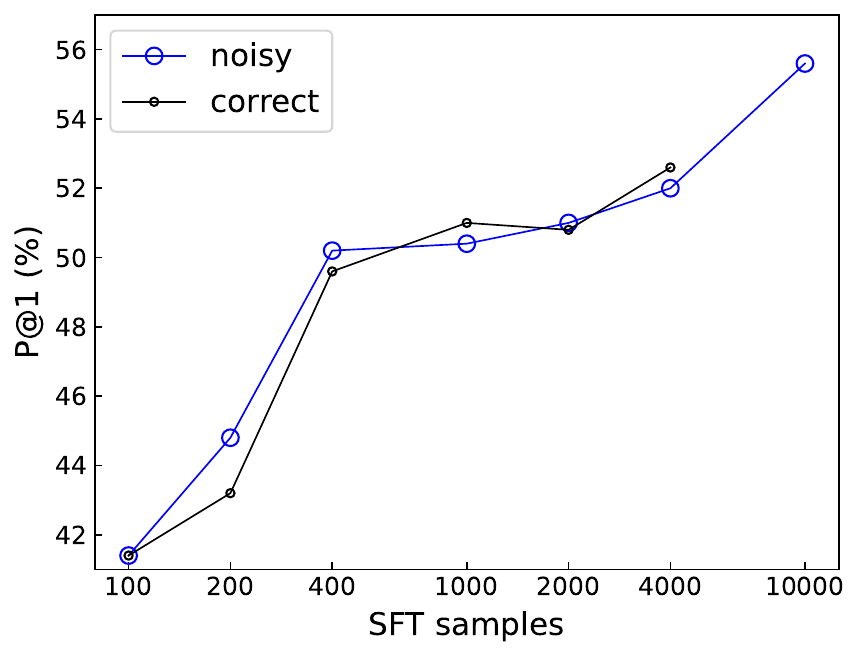}
\caption{All generated instructions (noisy) vs correct instructions only on MBPP. The correctness is verified by test cases generated from gpt-3.5-turbo-instruct. Using noisy instructions does not cause a significant performance decline.}
\label{fig:all_correct}
\end{figure}

We take MBPP as the object and examine how a decline in instruction quality affects the results. Specifically, we analyze the impact of using potentially erroneous instructions generated by Ada-Instruct (denoted as noisy samples) compared to using correct instructions. To determine the correctness of the instructions, given that MBPP samples include both code and use cases, we test whether the generated code passes through these cases successfully. Instructions that do so are considered correct samples. Among all noisy samples generated, we found that 46.9\% are correct. We sampled different scales of generated noisy samples and correct samples, respectively, and compared the effects of training models on them in Figure~\ref{fig:all_correct}.

We observed that the effects on the originally generated noisy samples are comparable to those based on correct samples, echoing a similar finding in~\cite{honovich2022unnatural}. This indicates that the difference in effectiveness between noisy samples produced by open-source LLMs and those produced by closed-source LLMs might not be a significant concern in sample generation. Even for complex tasks like programming, the impact of using noisy instructions generated by Ada-Instruct appears to be minimal. This confirms Ada-Instruct's adaptability in handling instructional noise.


\subsection{Scaling Up the Instructions}

\begin{figure}[tb]\label{fi}
\centering
\includegraphics[width=0.42\textwidth]{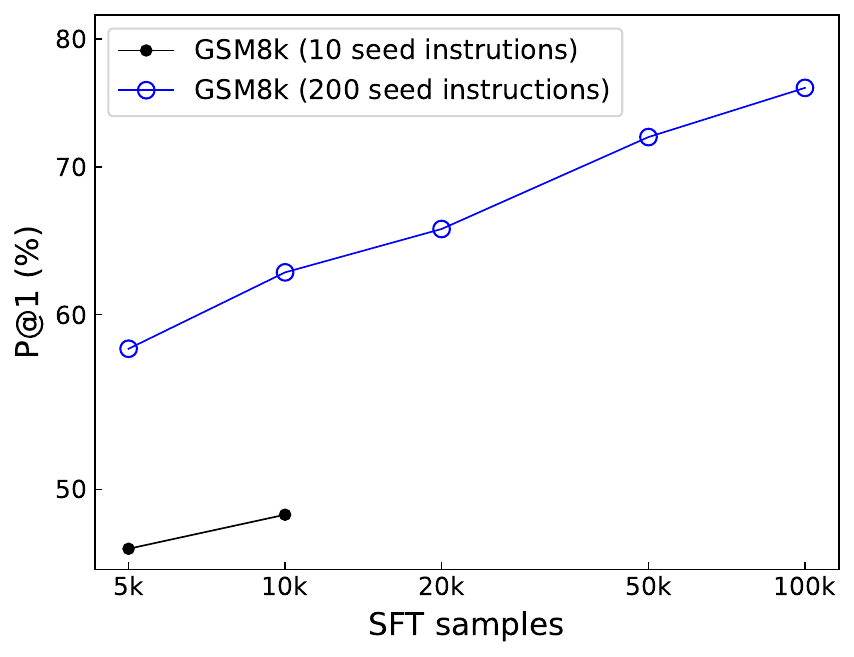}
\caption{Impact of increasing both the number of seed samples and the number of SFT samples. Both $x$ and $y$ axes are presented on a log scale.}
\label{exp:scaling}
\end{figure}

We further validate the efficacy of Ada-Instruct by increasing both the number of seed samples (for example, 200 seed instructions) and the scale of SFT samples. Figure~\ref{exp:scaling} illustrates our experimental results on GSM8k. A larger set of seed instructions leads to improved performance. Under the condition of 200 seed instructions, the P@1 and the number of SFT samples exhibit a clear scaling law, with room for further improvement. This evidence substantiates that Ada-Instruct's performance significantly improves as the instruction size increases.

\section{Conclusion}

We unveil novel insights into the capabilities of instruction generation, demonstrating that the conventional ICL-based Self-Instruct fails to generate long and complex instructions. In contrast, we reveal the proficiency of fine-tuning in generating task-aligned instructions, even with a limited number of initial samples. We introduced Ada-Instruct, a novel few-shot instruction generation methodology that leverages the fine-tuning of open-source LLMs, diverging significantly from the prevalent self-instruct strategies based on in-context learning with closed-source LLMs. Ada-Instruct ensures the generation of coherent, high-quality, and diverse instructions that align well with the target task distribution, presenting a groundbreaking solution to the challenges of data sparsity and diversity in instruction generation. 

\section{Limitations}

There are a few limitations worth noting:
\begin{itemize}
    \item Reliance on closed-source LLMs for labeling: In the current implementation of Ada-Instruct, the labeling step relies on a closed-source LLM (e.g. ChatGPT). The performance and reliability of the labeling step are subject to the capabilities and limitations of the chosen closed-source LLM.
    \item Limited evaluation on more tasks: The experiments in this paper primarily focus on code completion, mathematical reasoning, and commonsense reasoning tasks. Further evaluation on a wider range of tasks is helpful to comprehensively assess the generalizability and effectiveness of Ada-Instruct.
\end{itemize}

\bibliography{custom}

\appendix



\newpage
\appendix

\section{Quality Analysis}
\label{sec:exp:quality}

To assess the quality of the generated instructions, we evaluated whether the generated instructions are coherent and logically sound. For this evaluation, we used ChatGPT as an annotator. We randomly sampled 200 generated instructions for MBPP and CommonsenseQA. We first tell ChatGPT the task description of MBPP and CommonsenseQA, and then ask ChatGPT, ``Do you think this instruction is coherent and logically sound? Yes or No.'' As a baseline, we also evaluated the quality of the real samples from the corresponding data sets as the upper quality limit.  

As can be seen in Table~\ref{tab:quality}, the quality of the generated instructions is comparable to that of the real samples, suggesting that the generated samples possess sufficient accuracy. 
Although a small fraction of incorrect samples still exist, we investigated the impact of such errors in Section~\ref{sec:exp:qualityaffect}.

\begin{table}[h]
\centering
\fontsize{8pt}{10pt}\selectfont
\setlength{\tabcolsep}{2pt}
\begin{tabularx}{\columnwidth}{ccc|ccc}
\toprule
\multicolumn{3}{c|}{MBPP}      & \multicolumn{3}{c}{CommonsenseQA} \\
Generated   & Real Samples & Ratio  & Generated    & Real Samples  & Ratio   \\ \midrule
80.5\% & 93.0\%       & 86.6\% & 62.0\%  & 65.0\%        & 95.4\% \\
\bottomrule
\end{tabularx}
\caption{Quality of generated instructions, evaluated by ChatGPT. We compare with the real instructions, showing that their quality are close.}
\label{tab:quality}
\end{table}

\section{Impact of Length on Performance}
Ada-Instruct's ability to generate longer instructions that align well with the target distribution contributes to its performance improvement. To directly validate the benefits of longer instructions experimentally, we selected HumanEval as the target task. We randomly sampled two sets of 5k instructions:
\begin{enumerate}
    \item From all instructions generated by Ada-Instruct.
    \item Only from instructions with lengths less than 90 (based on Figure~\ref{fig:length}, self-instruct rarely generates instructions longer than 90 tokens).
\end{enumerate}

\begin{table}[h]
\centering
\begin{tabularx}{\columnwidth}{p{70pt}<{\raggedright}|>{\centering\arraybackslash}X}
\toprule
Length & HumanEval \\
\midrule
Length $<$ 90 & 57.9 \\
Full Length & 61.0 \\
\bottomrule
\end{tabularx}
\caption{Comparison of pass@1 (\%) results on HumanEval using two distinct sets of 5k instructions.}
\label{tab:length}
\end{table}

As shown in Table~\ref{tab:length}, instructions sampled from the set that includes longer examples yield a higher pass@1 score.

\section{Training Details}
When fine-tuning in Step 1, we train the models for 40 epochs with 10\% warm-up steps for all tasks. We use a batch size of 10, a learning rate of 1e-6, a weight decay of 1e-2, a cosine learning rate scheduler, and bf16 precision for all tasks except for MATH. We find MATH much harder than other tasks, so we apply a lower learning rate of 8e-7 to better adapt to the task. For all tasks under consideration, we adopt the first checkpoint at which the loss value resides within the range of 0.2 to 0.4 to avoid overfitting. This checkpoint is selected from the 25th, 30th, 35th, and 40th training epochs.

In Step 1 of the generation process, the temperature is set to $1$ for all tasks. To enhance diversity, we utilized top-k sampling. Specifically, for simpler MBPP and CSQA, we set $k=100$, while for more complex HumanEval, GSM8K, and MATH, we set $k=80$.

When fine-tuning in Step 3, for all tasks except HumanEval and CommonsenseQA, we train the LLMs for 3 epochs with a batch size of 256, a learning rate of 2e-5, a weight decay of 1e-2 and bf16 precision. We use a cosine scheduler with 10\% warm-up steps. For HumanEval, we adopt a lower learning rate of 1e-5. For CommonsenseQA, we adopt 2 training epochs and a lower learning rate of 1e-5, given that the data points in this task are much shorter than those in other tasks. Similarly to \cite{roziere2023code}, we adopt a cosine scheduler with 15\% warm-up steps and set the final learning rate to be 25\% of the peak learning rate. We do not apply loss masking to the instruction for all tasks except for CommonsenseQA, as the output for CommonsenseQA consists of only a few tokens.

\begin{table*}[tb]
\centering
\begin{tabularx}{\textwidth}{lX}
\toprule
\textbf{Model} & \textbf{Instruction} \\
\midrule
Self-Instruct & Given a list of words, create a dictionary to count the number of occurrences of each word. \\
\midrule
Evol-Instruct & Create a program that can filter out words of a string that contain a specific character and have a length greater than 3. Additionally, if the character is a vowel, the program should replace it with the next vowel in the vowel sequence. The program should then output the modified string, while maintaining the original word order. \newline
Additionally, you need to handle cases where the string contains special characters or numbers. If a word contains any special characters or numbers, it should be excluded from the output. \\
\midrule
{\bf Ada-Instruct} & You are given an array of meeting time ranges in any order. Each meeting time ranges[i] = [start\_i, end\_i] means that you need attend a meeting during the time range [start\_i, end\_i). Return the minimum number of conference rooms required. \\\bottomrule
\end{tabularx}
\caption{Comparison of Generated Instructions for HumanEval: Instructions from Self-Instruct are overly simplistic. Instructions from Evol-Instruct, while longer, exhibit unnaturalness and lack alignment with the target distribution. In contrast, Ada-Instruct successfully generates longer instructions that are consistent with the target distribution (algorithmic problems).}
\label{tab:case_study}
\end{table*}

\section{Case Study} 
In Table~\ref{tab:case_study}, we present the instructions generated by Ada-Instruct on HumanEval. We observe that the instructions generated by Self-Instruct are predominantly short. Although Evol-Instruct can generate longer instructions by iteratively adding constraints, these instructions tend to be unnatural and do not align well with the distribution of the downstream tasks. In contrast, Ada-Instruct is capable of producing longer instructions that align well with the target task. 


\section{Licenses for Artifacts}
We list the artifacts used in this paper and their licenses below:
\begin{itemize}
    \item \cite{touvron2023llama}, llama2
    \item \cite{xu2023wizardlm,luo2023wizardmath,luo2023wizardcoder}, llama2
    \item \cite{wang2022self}, Apache-2.0 license
\end{itemize}

This work is consistent with their intended use.

\section{Evaluation Strategies}

\subsection{Prompts for Downstream Tasks}
{\bf HumanEval:}
\begin{tcolorbox}[breakable]
\raggedright
\texttt{[INST] You are an expert Python programmer, complete the function below based on its docstring and the given test cases: \\
\{Question\} \\
Your code should start with a [PYTHON] tag and end with a [/PYTHON] tag. [/INST]}
\end{tcolorbox}

\noindent{\bf MBPP:}
\begin{tcolorbox}[breakable]
\raggedright
\texttt{[INST] You are an expert Python programmer, and here is your task: \{Question\} \newline
Your code should pass these tests: \newline
\newline
\{Test Cases\} \newline
Your code should start with a [PYTHON] tag and end with a [/PYTHON] tag. [/INST]}
\end{tcolorbox}

\noindent{\bf GSM8k and MATH:}
\begin{tcolorbox}[breakable]
\raggedright
\texttt{[INST] You are expert at solving math problems that require multi-step reasoning, and here is your task:\\
\{Question\} [/INST] Let's think step by step. \\
}
\end{tcolorbox}

\noindent{\bf CommonsenseQA:}
\begin{tcolorbox}[breakable]
\raggedright
\texttt{[INST] You are expert at commonsense reasoning, and here is your task: \{Question\} \\
A. \{Text of Label A\} \\
B. \{Text of Label B\} \\
C. \{Text of Label C\} \\
D. \{Text of Label D\} \\
E. \{Text of Label E\} [/INST] The answer is:}
\end{tcolorbox}

\subsection{Decoding Strategies}
For code completion tasks, to ensure comparable evaluations, we follow \cite{roziere2023code} and report the pass@1 scores of our models within the settings of greedy decoding and zero-shot.

For math tasks, to ensure comparable evaluations, we follow \cite{luo2023wizardmath} and report the pass@1 scores of our models within the settings of greedy decoding, zero-shot, and chain-of-thought.

For CommonsenseQA, the absence of an available test set necessitates the evaluation of our model on the development set. This evaluation is carried out within a framework adapted from \cite{hendrycks2020measuring}, and is executed in a zero-shot and answer-only manner. To ensure an equitable comparison, we also evaluate other LLAMA 2 base models in this setting.

\section{Fine-Tuning Data Formats for Ada-Instruct}
\subsection{Step 1}\label{sec:appendix:step1}
{\bf HumanEval:}
\begin{tcolorbox}[breakable]
\raggedright
\texttt{[INST] You are an expert Python programmer, complete the function below based on its docstring and the given test cases: \\
\{Question\} \\
Your code should start with a [PYTHON] tag and end with a [/PYTHON] tag. [/INST] [PYTHON] \\
\# pass \\}
\texttt{[/PYTHON]}
\end{tcolorbox}

\noindent{\bf MBPP:}
\begin{tcolorbox}[breakable]
\raggedright
\texttt{[INST] You are an expert Python programmer, and here is your task: \{Question\} \newline
Your code should pass these tests: \newline\newline
\{Test Cases\} \newline
Your code should start with a [PYTHON] tag and end with a [/PYTHON] tag. [/INST] [PYTHON] \newline
\# pass \newline}
\texttt{[/PYTHON]}
\end{tcolorbox}

\noindent{\bf GSM8k and MATH:}
\begin{tcolorbox}[breakable]
\raggedright
\texttt{[INST] You are expert at solving math problems that require multi-step reasoning, and here is your task:\\
\{Question\} [/INST] Let's think step by step. \\
}
\end{tcolorbox}

\noindent{\bf CommonsenseQA:}
\begin{tcolorbox}[breakable]
\raggedright
\texttt{[INST] You are expert at commonsense reasoning, and here is your task: \{Question\} \\
A. \{Text of Label A\} \\
B. \{Text of Label B\} \\
C. \{Text of Label C\} \\
D. \{Text of Label D\} \\
E. \{Text of Label E\} [/INST] }
\end{tcolorbox}

\subsection{Step 3}
{\bf HumanEval:}
\begin{tcolorbox}[breakable]
\raggedright
\texttt{[INST] You are an expert Python programmer, complete the function below based on its docstring and the given test cases: \\
\{Question\} \\
Your code should start with a [PYTHON] tag and end with a [/PYTHON] tag. [/INST] [PYTHON] \\
\{Output\} \\}
\texttt{[/PYTHON]}
\end{tcolorbox}

\noindent{\bf MBPP:}
\begin{tcolorbox}[breakable]
\raggedright
\texttt{[INST] You are an expert Python programmer, and here is your task: \{Question\} \newline
Your code should pass these tests: \newline\newline
\{Test Cases\} \newline
Your code should start with a [PYTHON] tag and end with a [/PYTHON] tag. [/INST] [PYTHON] \newline
\{Output\} \newline}
\texttt{[/PYTHON]}
\end{tcolorbox}

\noindent{\bf GSM8k and MATH:}
\begin{tcolorbox}[breakable]
\raggedright
\texttt{[INST] You are expert at solving math problems that require multi-step reasoning, and here is your task:\\
\{Question\} [/INST] Let's think step by step. \\
\{Output\}
}
\end{tcolorbox}

\noindent{\bf CommonsenseQA:}
\begin{tcolorbox}[breakable]
\raggedright
\texttt{[INST] You are expert at commonsense reasoning, and here is your task: \{Question\} \\
A. \{Text of Label A\} \\
B. \{Text of Label B\} \\
C. \{Text of Label C\} \\
D. \{Text of Label D\} \\
E. \{Text of Label E\} [/INST] The answer is: \{Output\}}
\end{tcolorbox}

\section{Prompts for Self-Instruct}
To encourage the generation of high quality and diverse instruction, we use the following prompts in the Self-Instruct baseline.

\subsection{Prompts For gpt-3.5-turbo-instruct}

\noindent{\bf HumanEval:}
\begin{tcolorbox}[breakable]
\raggedright
\texttt{You are asked to come up with a set of 20 diverse instructions on code completion task. These instructions will be given to a Codex model and we will evaluate the Codex model for generating codes that follow the instructions. \newline\newline
Here are the requirements: \\
1. The instructions are designed for testing the Python programming capability to solve Python problems. Each instruction should describe a Python problem with function definition, docstring, and test cases. \\
2. The instructions should incorporate as many Python concepts as possible, as well as being diverse and comprehensive. \\
3. The instructions should not be too easy. Each Python problem should be solved using built-in libraries or data structures with algorithm of intermediate level. \\
4. The instructions should at least 1 to 2 sentences long. Either an imperative sentence or a question is permitted. \\
5. The output should be an appropriate response to the instruction, and should take full account of requirements and test cases in the instruction. \\
6. The instructions must not appear in mainstream evaluation datasets for code generation, e.g. HumanEval, MBPP, DS1000 and so on. \newline\newline
List of 20 tasks: \newline
\#\#\# \newline
1. \{Example 1\} \\
\#\#\# \newline
2. \{Example 2\} \\
\#\#\# \newline
3. \{Example 3\} \newline
\#\#\# \newline
4. }
\end{tcolorbox}

\noindent{\bf MBPP:}
\begin{tcolorbox}[breakable]
\raggedright
\texttt{You are asked to come up with a set of 20 diverse instructions on code completion task. These instructions will be given to a Codex model and we will evaluate the Codex model for generating codes that follow the instructions. \newline\newline
Here are the requirements: \\
1. The instructions are designed for testing the Python programming capability to solve basic Python problems. Each instruction should have a clear and distinct solution. \\
2. The instructions should incorporate as many Python concepts as possible, as well as being diverse and comprehensive. \\
3. The instructions should not be too complicated or too easy. Each Python problem should be solved using built-in libraries or data structures with algorithm of 
intermediate level. \\
4. The instructions should at least 1 to 2 sentences long. Either an imperative sentence or a question is permitted. \\
5. The output should be an appropriate response to the instruction, and should take full account of requirements and test cases in the instruction. \\
6. The instructions must not appear in mainstream evaluation datasets for code generation, e.g. HumanEval, MBPP, DS1000 and so on. \newline\newline
List of 20 tasks: \\
\#\#\# \\
1. \{Example 1\} \\
\#\#\# \\
2. \{Example 2\} \\
\#\#\# \\
3. \{Example 3\} \\
\#\#\# \\
4. }
\end{tcolorbox}

\noindent{\bf GSM8k:}
\begin{tcolorbox}[breakable]
\raggedright
\texttt{You are asked to come up with a set of 20 diverse instructions on math problem solving task. These instructions will be given to a math model and we will evaluate the math model for generating solutions that follow the instructions. \newline\newline
Here are the requirements: \\
1. The instructions are designed for testing the math capability to solve math problems that require multi-step reasoning. Each instruction should be accompanied by a detailed reasoning path and a final answer. \\
2. The instructions should include diverse types of grade school math problems, as well as being diverse and comprehensive. \\
3. The instructions should not be too complicated or too easy. Each math problem should take between 2 and 8 steps to solve, and solutions primarily involve performing calculations using basic arithmetic operations (+ - / *) to reach the final answer. \\
4. The instructions should at least 1 to 2 sentences long. Either an imperative sentence or a question is permitted. \\
5. The output should be an appropriate response to the instruction that is in the form of reasoning followed by the final answer. \\
6. The instructions must not appear in mainstream evaluation datasets for math, e.g. GSM8K, MATH and so on. \newline\newline
List of 20 tasks: \\
\#\#\# \\
1. \{Example 1\} \\
\#\#\# \\
2. \{Example 2\} \\
\#\#\# \\
3. \{Example 3\} \\
\#\#\# \\
4. }
\end{tcolorbox}

\noindent{\bf MATH:}
\begin{tcolorbox}[breakable]
\raggedright
\texttt{You are asked to come up with a set of 20 diverse instructions on math problem solving task. These instructions will be given to a math model and we will evaluate the math model for generating solutions that follow the instructions. \newline\newline
Here are the requirements: \\
1. The instructions are designed for testing the math capability to solve math problems that require multi-step reasoning. Each instruction should be accompanied by a detailed reasoning path and a final answer. \\
2. The instructions should describe math problems in LaTex that require knowledge such as calculus, algebra, number theory, counting and probability, etc. \\
3. The instructions should be challenging, diverse and comprehensive. Each math problem should take multiple steps of complex reasoning maybe with some advanced mathematical knowledge and tools to solve. \\
4. The instructions should at least 1 to 2 sentences long. Either an imperative sentence or a question is permitted. \\
5. The output should be an appropriate response to the instruction that is in the form of reasoning followed by the final answer. Both the reasoning and answer should be in the form of LaTex. The final answer should be placed in "\$\textbackslash boxed\{\}\$". \\
6. The instructions must not appear in mainstream evaluation datasets for math, e.g. GSM8K, MATH and so on. \newline\newline
List of 20 tasks: \\
\#\#\# \\
1. \{Example 1\} \\
\#\#\# \\
2. \{Example 2\} \\
\#\#\# \\
3. \{Example 3\} \\
\#\#\# \\
4. }
\end{tcolorbox}

\subsection{Prompts For gpt-4o}
\noindent{\bf HumanEval:}
\begin{tcolorbox}[breakable]
\raggedright
\texttt{\textbf{user:} You are asked to come up with a set of 10 diverse instructions on code completion task. These instructions will be given to a Codex model and we will evaluate the Codex model for generating codes that follow the instructions. \newline\newline
Here are the requirements: \\
1. The instructions are designed for testing the Python programming capability to solve Python problems. Each instruction should describe a Python problem with function definition, docstring, and test cases. \\
2. The instructions should incorporate as many Python concepts as possible, as well as being diverse and comprehensive. \\
3. The instructions should not be too easy. Each Python problem should be solved using built-in libraries or data structures with algorithm of intermediate level. \\
4. The instructions should at least 1 to 2 sentences long. Either an imperative sentence or a question is permitted. \\
5. The output should be an appropriate response to the instruction, and should take full account of requirements and test cases in the instruction. \\
6. The instructions must not appear in mainstream evaluation datasets for code generation, e.g. HumanEval, MBPP, DS1000 and so on.\newline\newline
\textbf{assistant:} \#\#\# \\
1. \{Example 1\} \\
\#\#\# \\
2. \{Example 2\} \\
\#\#\# \\
3. \{Example 3\} \\
\#\#\# \newline\newline
\textbf{user:} Continue to generate the remaining 7 instructions. The order number of each instruction must be preceded by "\#\#\#".
}

\end{tcolorbox}

\noindent{\bf GSM8k:}
\begin{tcolorbox}[breakable]
\raggedright
\texttt{\textbf{user:} You are asked to come up with a set of 10 diverse instructions on math problem solving task. These instructions will be given to a math model and we will evaluate the math model for generating solutions that follow the instructions. \newline\newline
Here are the requirements: \\
1. The instructions are designed for testing the math capability to solve math problems that require multi-step reasoning. Each instruction should be accompanied by a detailed reasoning path and a final answer. \\
2. The instructions should include diverse types of grade school math problems, as well as being diverse and comprehensive. \\
3. The instructions should not be too complicated or too easy. Each math problem should take between 2 and 8 steps to solve, and solutions primarily involve performing calculations using basic arithmetic operations (+ - / *) to reach the final answer. \\
4. The instructions should at least 1 to 2 sentences long. Either an imperative sentence or a question is permitted. \\
5. The output should be an appropriate response to the instruction that is in the form of reasoning followed by the final answer. \\
6. The instructions must not appear in mainstream evaluation datasets for math, e.g. GSM8K, MATH and so on.\newline\newline
\textbf{assistant:} \#\#\# \\
1. \{Example 1\} \\
\#\#\# \\
2. \{Example 2\} \\
\#\#\# \\
3. \{Example 3\} \\
\#\#\# \newline\newline
\textbf{user:} Continue to generate the remaining 7 instructions. The order number of each instruction must be preceded by "\#\#\#".
}

\end{tcolorbox}

\end{document}